\documentclass[review]{elsarticle}

\usepackage{hyperref}

\usepackage{xcolor}
\usepackage{textcomp}
\usepackage{amsmath,amssymb,amsfonts}

\usepackage{lineno,hyperref}
\modulolinenumbers[5]

\journal{Journal of \LaTeX\ Templates}

\begin{document}

\begin{frontmatter}

\title{Hierarchical Complementary Learning for Weakly Supervised Object Localization}
\tnotetext[mytitlenote]{Fully documented templates are available in the elsarticle package on \href{http://www.ctan.org/tex-archive/macros/latex/contrib/elsarticle}{CTAN}.}

\author[mymainaddress]{Sabrina Narimene Benassou}
\ead{benassou.narimene@hit.edu.cn}

\author[mysecondaryaddress]{Wuzhen Shi}
\ead{wzhshi@szu.edu.cn}

\author[mymainaddress]{Feng Jiang\corref{mycorrespondingauthor}}




\cortext[mycorrespondingauthor]{Corresponding author}
\ead{fjiang@hit.edu.cn}

\author[mythirdaddress]{Abdallah Benzine}
\ead{Abdallah.benzine@digeiz.com}

\address[mymainaddress]{Harbin Institute of Technology, 92 Xidazhi Street, Nangang District, Harbin City, Heilongjiang Province, China}
\address[mysecondaryaddress]{College of Electronics and Information Engineering, Shenzhen University No. 3688, Nanhai Avenue, Nanshan District, Shenzhen, China}
\address[mythirdaddress]{Digeiz, AI lab, 47 RUE MARCEL DASSAULT 92100 BOULOGNE-BILLANCOURT, France}

\begin{abstract}
Weakly supervised object localization (WSOL) is a challenging problem which aims to localize objects with only image-level labels. Due to the lack of ground truth bounding boxes, class labels are mainly employed to train the model. This model generates a class activation map (CAM) which activates the most discriminate features. However, the main drawback of CAM is the ability to detect just a part of the object. To solve this problem, some researchers have removed parts from the detected object \cite{b1, b2, b4}, or the image \cite{b3}. The aim of removing parts from image or detected parts of the object is to force the model to detect the other features. However, these methods require one or many hyper-parameters to erase the appropriate pixels on the image, which could involve a loss of information. In contrast, this paper proposes a Hierarchical Complementary Learning Network method (HCLNet) that helps the CNN to perform better classification and localization of objects on the images. HCLNet uses a complementary map to force the network to detect the other parts of the object. Unlike previous works, this method does not need any extras hyper-parameters to generate different CAMs, as well as does not introduce a big loss of information. In order to fuse these different maps, two different fusion strategies known as the addition strategy and the l1-norm strategy have been used. These strategies allowed to detect the whole object while excluding the background.  Extensive experiments show that HCLNet obtains better performance than state-of-the-art methods.
\end{abstract}


\end{frontmatter}


\section{Introduction}
\label{sec:introduction}
Convolutional Neural Network has attracted many researchers around the world and has become the first tool to use for object detection, object localization, face recognition, image classification, etc. However, in order to achieve these tasks, 
CNNs require a lot of annotated data which may be difficult to obtain. Even if getting class labels for natural images nowadays is getting easier due to some smart annotation tools \cite{b15}. Nevertheless, getting bounding box annotations is more problematic for object localization. The need of training CNNs without ground truth bounding boxes is becoming crucial, especially for some area such as medical image processing, where an expert is needed to annotate images.

\begin{figure}[t]
\begin{minipage}[t]{1.0\linewidth}
  \centering
  \centerline{\includegraphics[width=8cm, height=9cm]{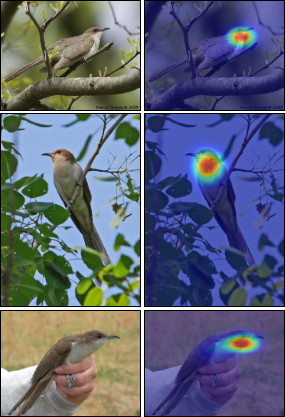}}
\caption{Visualization of CAM \cite{b2}. Only the most discriminative part is highlighted by the CAM i.e the head of the birds, but the body is not. }
\label{fig:cam}
\end{minipage}
\end{figure}

Weakly Supervised Learning (WSL) aims at solving this problem. A method has been highly used to localize objects on the image known as class activation map (CAM) \cite{b2}. However, CAM is able to localize only the most discriminative part of an object and does not highlight the whole object \cite{b2}. As shown in Fig. \ref{fig:cam}, the heads of birds are the features that discriminate the object and then highlighted on the map. On the other hand, the body is not discriminative enough for the model to be highlighted.

To avoid the problem of detecting only a part of the object, many methods have been proposed. These methods can be divided into two approaches. The first approach follows the Multiple Instance Learning (MIL) pipeline, where images are considered as bags, and region proposals generated by objects proposals techniques as instances \cite{b22,b23}. Unfortunately, the CNN selects the proposal with the highest score. This proposal does not cover the whole object which result in a poor localization accuracy. The second approach is to remove one part or some parts of the CAM. For example \cite{b2,b3,b4} remove the most discriminative part, so that the CNN is forced to detect the other part of the object. However, erasing the most discriminative part reduces the recognition performance and does not show significant localization accuracy improvement. Furthermore, zeroed-out this part implies non informative pixels for the network. The non-informative pixels are considered as a conceptual limitation for CNNs as they are data hungry \cite{b25}. In addition, these methods need an extra hyper-parameter i.e. a threshold to consider which pixels should be zeroed-out. This threshold is not easy to determine due to its variation for each dataset and model used for training.

In this paper, we follow Adversarial Complementary Object Localization (ACoL) to generate our CAMs \cite{b1}. To the best of our knowledge no work has reported a complementary CAM (C-CAM) to ensure that there is no degradation in the classification accuracy without the use of any threshold. The C-CAM uses complement values of CAM instead of using a threshold to erase the appropriate pixels and hence the discriminative part is not zeroed-out. The C-CAM can be used regardless of the model or the dataset that we decide to employ. In the following contexts, the C-CAM shows an improvement in classification and localization accuracy. 

In order to get the final location of the object, the generated CAMs need to be fused, Zhang et al. \cite{b1} fuse them by taking the maximum values. ACoL took the maximum of each CAMs to construct the final map. However, the final CAM highlights the background and not only the foreground. To solve this problem, we introduce two different fusion strategies, the addition strategy and the $l_{1}$-norm strategy \cite{b7}. These two methods take into consideration both values in the maps instead of one value. The addition strategy sum both features detected by the two branches. However, it is a severe strategy compared to $l_{1}$-norm. To represent the importance of each region of the map, the $l_{1}$-norm multiplies each pixel by its weight before to add up the features together.

The contributions of this paper include three aspects: 
\begin{enumerate}
\item A hierarchical Complementary learning network (dubbed HCLNet) is proposed for weakly supervised object localization. HCLNet hierarchically generates different class activation maps, and fuses them to effectively locate the whole object.
\item Two different fusion strategies, the addition strategy and the $l_{1}$-norm strategy, have been introduced to fuse these CAMs and make it able to detect the whole object. 
\item Extensive experiments to show that the proposed method achieves a new state-of-the-art performance for weakly supervised localization
\end{enumerate}

\section{Related work}
\label{sec:related work}
Weakly supervised learning has became an exciting subject due to the lack of bounding boxes annotations. Many works have proposed different techniques to localize objects using only image-level labels. In this section, we introduce the state of the art methods for weakly supervised object localization and weakly supervised object detection, as well as briefly review some feature fusion methods.
\subsection{Weakly Supervised Object Localization}
Weakly Supervised Object Localization aims to localize objects by using only their class label. In \cite{b2}, Global Average Pooling (GAP) is added to the CNN, each unit in the GAP is multiplied by its corresponding feature map and then summed up together to generate the Class Activation Map. CAM highlights the most discriminative part of the object, but it cannot detect the whole object. Some methods deal with this problem as in \cite{b1} by using two branches. The first branch takes as input the full image to detect the most discriminative part and removes it. Then the second branch takes the output of the first branch to detect the other part of the object. \cite{b3} is another method that uses a hiding process where each image is divided into patches. For each patch, a probability is assigned. At each epoch some patches are hided and given as input to the CNN. The model then learns to detect the whole object instead of a small part of the object. \cite{b4} has added a module after some blocks of the network, where a dropout is used to hide the most discriminative part detected by the feature map and an attention mechanism to reward it, each component is used stochastically by the model to learn the whole features of the object. In \cite{b5}, the CNN generates an attention map that highlights the most discriminative region of the object. This attention map guides the model to detect the other part of the object by using two thresholds. One threshold is to decide the foreground pixels, the second threshold is to decide the background pixels. All the cited methods propose to erase a part or some parts of the image in order to force the network to localize the full object. But to remove these parts a threshold is needed which is not easy to determine.

\subsection{Weakly Supervised Object Detection}
In Weakly Supervised Object Detection (WSOD) several works attempt to use Multiple Instance Learning (MIL) pipeline. MIL consists of considering the image as a bag and region proposals as instances. In \cite{b8}, some proposals cover the object but do not fill the requirement IOU $>$ 50\%. The region proposal which has the highest score is selected, then the returned bounding box partially covers the object. Some works use different techniques to select a cluster of proposals. In \cite{b9}, the model refines the selected region proposal by adding online instance classifier refinement branches. Online instance classifier refinement branches take into account the proposals that overlap the highest one, then the CNN detects more parts of the object. \cite{b14}, proceeds in a cascaded way and consists in three stages. The first stage consists of a fully convolutional network that generates a CAM used by the second stage. The second stage performs a class specific segmentation to extract proposals. Finally the third stage performs MIL on the generated proposals. \cite{b10}, follows \cite{b9} but instead of taking all the proposals to perform an MIL, they select the proposals according to their context. \cite{b13} minimizes the global entropy to select the high-scored proposals followed by the minimization of the local entropy to identify the correct proposals. In \cite{b11}, rather than adding a regularizer into the loss function, an optimization method called continuation method is used to smooth the loss function by deviding the problem into multiple sub-problems. In \cite{b12}, two modules are used. The first module is Pseudo ground-truth excavation (PGE) that removes the most discriminative proposals and merges the remaining ones. The second module is Pseudo ground-truth adaptation (PGA) which trains a region proposal network that generates the final pseudo ground-truth bounding boxes. Then, the final ground-truth bounding boxes are used by the CNN for training.

\subsection{Feature fusion}
Feature maps from different layers in the CNN hold different information, fusing these feature maps could provide better results. Different fusion techniques for different purposes such as object detection \cite{b16} or infrared and visible image \cite{b7} were proposed. Some works for instance have used: addition fusion and $l_{1}$-norm fusion strategies \cite{b7}, max fusion \cite{b1}, conv fusion and bilinear fusion \cite{b6}. In \cite{b6}, they combine the two-stream of ConvNets by applying 3D Conv to the convolutional layer then applying 3D Pooling. In \cite{b16} the authors combine the features from different level layers by concatenating them and feed them to the next layer to improve the classification and detection of the network. Herein, we propose to use addition fusion and $l_{1}$-norm fusion strategies to combine different parts of an object, and then improve the performance in weakly supervised object localization.

\section{The proposed method}
\label{sec:method}
\begin{figure*}[t]
\begin{minipage}[t]{1.0\linewidth}
  \centering
  \centerline{\includegraphics[width=15cm]{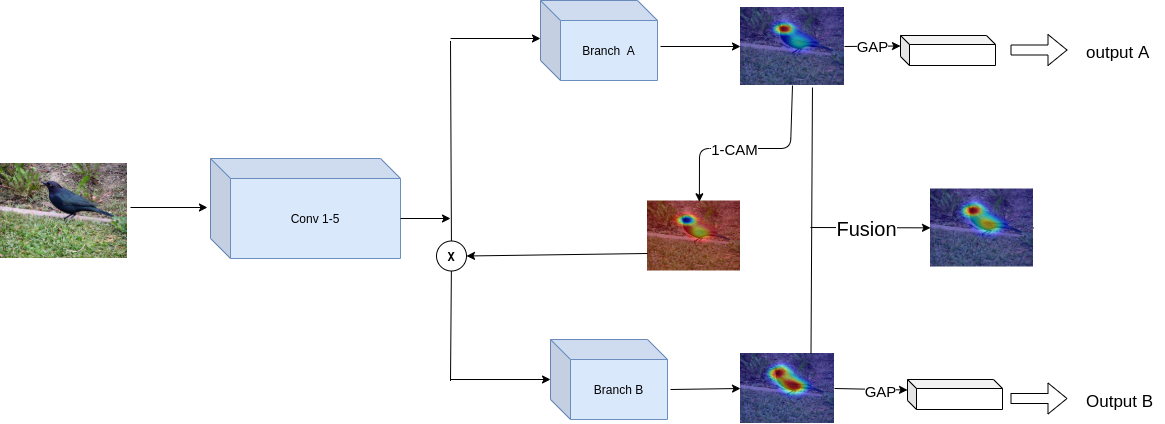}}
\caption{The proposed architecture. The full image is given to the network, The first branch generates the CAM that activate the most discriminate part. We apply our C-CAM and feed it to the second branch to generate the second map with the other part of the object highlighted. The CAMs are then fused using one of the two proposed strategies.}
\label{fig:met}
\end{minipage}
\end{figure*}

 In this section, we firstly review the state of the art ACoL, then present our complementary CAM (C-CAM) and finally introduce two fusion methods used in this article. The proposed architecture is reflected in Fig. \ref{fig:met}.

\subsection{Review of ACoL : Adversarial Complementary Learning for Weakly Supervised Object Localization}
ACoL consists of removing the layers after conv5-3 (from pool5 to prob) of VGG-16 network and adding two branches to the CNN. The image is given as input to the first branch (bracnh A). Branch A is responsible for generating the CAM with the most discriminative part of the object highlighted on the image. The most discriminative part is erased according to a threshold. The second branch (branch B) takes as input the thresholded CAM to generate a second CAM, where the other part of the object is erased. In the end, the two CAMs are fused by taking the maximum value of each map. More details are discussed in \cite{b1}.

\subsection{Complementary CAM (C-CAM)}
The main drawback of ACoL \cite{b1} is, it removes the most discriminative part according to a threshold. This threshold is an hyper-parameter that should be fixed during training. Noting that this hyper-parameter is not easy to choose because its value differs for each dataset and each model. Furthermore, this erasing operation creates a map where a region contains non-informative pixels. These non-informative pixels create a limitation for the network which decrease the classification accuracy. 

In the present work, we propose to use $1-CAM$ map instead of the thresholded map to guide branch B to detect the other part of the object. The CAM contains normalized pixels values between $0 < p^{(x,y)} < 1$. Subtracting 1 by the CAM, we ensure that there is still some information for the network. In addition, the C-CAM unlike previous methods does not require any hyperparameter. The following contexts show that our modification significantly improves the classifiction and the localization performance. 

\begin{figure}[t]
  \centering
  \centerline{\includegraphics[width=0.5\textwidth]{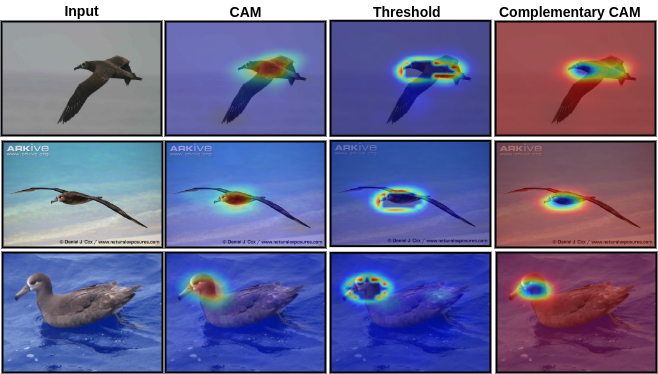}}
\caption{The difference between the thresholded CAM and the complementary CAM}
\label{fig:dif}  
\end{figure}

The differences between the maps are reflected in Fig \ref{fig:dif}. red color on the map refers to high pixels values, i.e. these pixels are more important for the network. The blue color refers to pixels with low values, i.e. they are less important. In the CAM the most discriminative part is red as it contains more information. The background and the body are blue as they contain less information. By creating a complementary map, we reverse the colors of the map, the most discriminative part becomes in blue and all the other pixels are in red. Thereby, we give the information to the second branch that it should focus on the other part of the object and not the most discriminative part. When using a threshold, some pixels of the most discriminative part are not removed from the map, then the second branch keep focusing on them during training.

\subsection{Feature fusion}
In ACoL \cite{b1}, both maps generated by the CNN are fused by taking the maximum value for each map. However, when taking the maximum value of each CAMs, the background is sometimes highlighted \cite{b1}. In addition, taking the maximun value favorises one value of the CAM over the other. In order to avoid these drawbacks, we propose to change this fusion method and instead to use two other methods \cite{b7} : addition strategy and $l_{1}$-norm strategy. 

\begin{figure}[t]
  \centering
  \centerline{\includegraphics[width=0.5\textwidth, height=5cm]{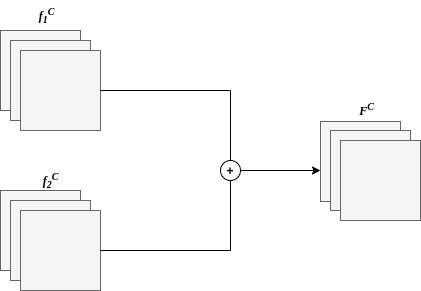}}
\caption{Addition of the two feature maps. Instead of concatenating the two feature maps, we sum up each feature map of the corresponding class together. }
\label{fig:add}  
\end{figure}

\textbf{Addition strategy}:
Let $f_1^C $ and $f_2^C $ be the set of feature maps of branch A and branch B respectively, where $c \in \{ 1, ... C \}$  and C is the number of classes. The addition strategy consists of summing up over each values of the two feature maps to get the final feature map $F^C$ (Eq \ref{eq:1}). The process is shown in Fig. \ref{fig:add}. 
\begin{eqnarray}
F^{C}(x,y) =  \sum_{i=1}^{2} f_{i}^{C}(x,y)
\label{eq:1}
\end{eqnarray}

$\textbf{L}_{\textbf{1}}$\textbf{-norm strategy}:
The addition strategy is severe as it sums both features detected by the two branches. $L_{1}$-norm prefers sparse weight vectors. This means that $l_{1}$-norm performs feature selection and excludes all features where the weight is 0. Thereby, it selects the pixels that belong to the object. 

First, $l_{1}$-norm is used to compute the activity level map using Eq. \ref{eq:2}. Then we apply the block-based average operator using Eq. \ref{eq:3} to calculate the final activity level map.

$f_i^C $ still denotes the set of feature maps, M is the primary activity level map and $\hat{M}_{i}$ is the final activity level map. 

\begin{eqnarray}
M_{i}(x,y) = {\|{f_{i}^{1:C}}(x,y)\|_{1}}
\label{eq:2}
\end{eqnarray}
\begin{eqnarray}
\hat{M}_{i}(x,y)=\frac{\sum_{\substack{a=-r}}^{r}\sum_{\substack{b=-r}}^{r}M_{i}(x + a, y + b)}{(2r+1)^{2}}
\label{eq:3}
\end{eqnarray}

As in \cite{b7}, r is the block size and is set to r = 1.

\begin{figure}[t]
  \centering
  \centerline{\includegraphics[width=0.5\textwidth]{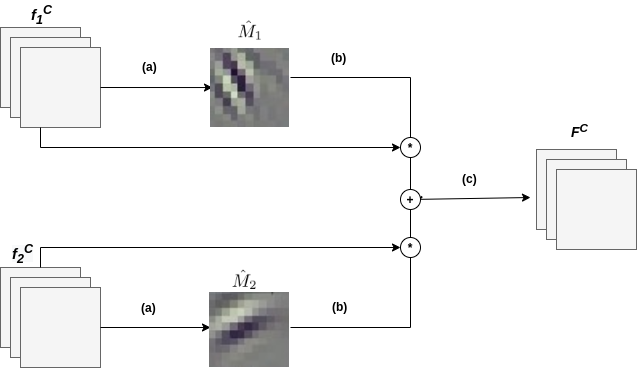}}
\caption{$l_{1}$-norm fusion method. (a) is the $l_{1}$-norm + block-based average operation that generates the activity level map. (b) calculation of the softmax of these two activity level maps to obtain a weight vector. (c) mutiplying each feature maps by its weight vetcor and then sum up the results. }
  
\label{fig:norm}
\end{figure}

We calculate the probability of each pixel of this activity level map by using the softmax function Eq. \ref{eq:5}, to obtain weights where a high value means an important pixel for the image. By multiplying each pixel by its weights Eq. \ref{eq:4}, we get a more informative map. The two generated informative maps are then sum up together to obtain the final map. The whole operations are depicted in Fig. \ref{fig:norm}.

\begin{eqnarray}
F^{C}(x,y)= \sum_{\substack{i=1}}^{2} w_{i}(x,y) \times f_{i}^{C}(x,y) 
\label{eq:4}
\end{eqnarray}

\begin{eqnarray}
where  \ w_{i}(x,y)=\frac{\hat{M}_{i}(x,y)}{\sum_{\substack{n=1}}^{2}\hat{M}_{n}(x,y)}
\label{eq:5}
\end{eqnarray}


\subsection{Loss Function}
The loss function is the same as in \cite{b1}, we apply the cross entropy on the results of the two branches and then add them up.
\begin{eqnarray}
\sum_{\substack{c=1}}^{C} y_{c} \log(p_{A,c}) + y_{c} \log(p_{B,c})
\end{eqnarray}

where \textit{C} is the number of classes, \textit{$y_{c}$} is the correct class of the image and $p_{A,c}$ and $p_{B,c}$ are the predicted probability class of the object by branch A and branch B, respectively.

\section{Experiments}
\label{sec:experiments}
\textbf{Datasets}: Two commonly datasets are used for WSOL evaluation: CUB-200-2011 \cite{b17} and ILSVRC 2016 \cite{b18,b20}. CUB-200-2011 contains 200 classes of bird species. The total are 11,788 images, with 5,994 images for training and 5,794 for test. ILSVRC 2016, is a large scale dataset of 1000 classes, that comprises 1.2 million for training, and 5,000 images for the validation set. The validation set is used for testing.

\textbf{Evaluation metrics}: We use the same evaluation metrics as in \cite{b1}. For classification, we use, Top-1 and Top-5 classification accuracy. The Top-1 (resp. Top-5) determines that the answer is correct when the class with the highest probability prediction (resp. the 5 highest probability predictions) is equal to the ground-truth class. For localization, Top-1 and Top-5 localization accuracy metrics are used. Top-1 loc (resp. Top-5 loc) determines that the localization is correct when the intersection over union (IOU) between the ground truth and the estimated top-1 (resp. top-5) bounding boxes is equal or greater than 50\%. We further use GT-known localization accuracy metric \cite{b3} also called Correct Localization (CorLoc) \cite{b24}. CorLoc determines that the localization is correct when the intersection over union (IOU) between the ground truth bouding box and the predicted bouding box is greather than 50\%,  whatever the classification is correct or not.

\textbf{Implementation details}: The contributions with VGGnet was evaluted \cite{b19}. The network was implemented using Py-Torch. The layers after conv5-3 (from pool5 to prob) of the VGG-16 network are removed. Then, two convolutional layers with kernel size 3 $\times$ 3, stride 1, pad 1 with 1024 units, and a convolutional layer of size 1 $\times$ 1, stride 1 with the number of units equal to the number of classes (200 units for CUB-200-2011 and 1000 units for ILSVRC) are added. VGG-16 is fine-tuned on the pre-trained weights of ILSVRC \cite{b18,b20}. The input images are randomly cropped to 224 $\times$ 224 pixels after being re-sized to 256 $\times$ 256 pixels. For classification, we average the scores from the softmax layer with 10 crops.

\subsection{Comparison with the state-of-the-art methods}
\label{sec:majhead}
HCLNet was compared with the state-of-the-art CAM \cite{b2}, ACoL \cite{b1}, SPG \cite{b5} and ADL \cite{b4} methods on the CUB-200-2011 test set. On ILSVRC valisation set, the method was compared to \cite{b2} and ACoL \cite{b1}. The results are reflected in Tab. \ref{tab:1} and Tab. \ref{tab:2} respectively. The CorLoc results are reflected in Tab. \ref{tab:corloc}.

\begin{table}[t]
\centering
 \begin{tabular}{|c|c|c|c|c|} 

 \hline
  &  \multicolumn{2}{c|}{cls-err} &  \multicolumn{2}{c|}{loc-err} \\ 
 \hline\hline
 Method & top1 & top5 & top1 & top5 \\ 
 \hline
 VGGnet-CAM \cite{b2} & 23.4 & 7.5 & 55.85 & 47.84 \\ 
 VGGnet-ACoL \cite{b1} & 28.1 & - & 54.08 & 43.49 \\
 VGGnet-SPG \cite{b5} & 24.5 & 7.9 & 51.07 & 42.15 \\
 VGGnet-ADL \cite{b4} & 34.73 & - & \textbf{47.64} & - \\
 
 HCLNet (addition) & \textbf{22.11} & \textbf{6.44} & 48.95 & \textbf{41.06} \\ 
 HCLNet ($l_{1}$-norm) & \textbf{22.11} & \textbf{6.44} & 49.07 & 41.16 \\

 \hline
 \end{tabular}
 \caption{Comparison of the state-of-the-art performance on the CUB-200-2011 test set.}
 \label{tab:1}
\end{table}

\begin{table}[t!]

\centering
 \begin{tabular}{|c|c|c|} 
 \hline
 Method & cls-err top1 & loc-err top1 \\ 
 \hline
 VGGnet-CAM \cite{b2} & 31,53 & 61.73 \\ 
 VGGnet-ACoL \cite{b1} & 33,27 & 62.57 \\
 
 HCLNet (addition) & \textbf{29,27} & \textbf{60.15} \\ 
 HCLNet ($l_{1}$-norm) & \textbf{29.27} & \textbf{60.15} \\

 \hline
 \end{tabular}
 \caption{Comparison of the state-of-the-art performance on the ILSVRC validation set.}
 \label{tab:2}
\end{table}

\begin{figure*}[ht]

  \centering
  \centerline{\includegraphics[scale=0.5, width=\textwidth]{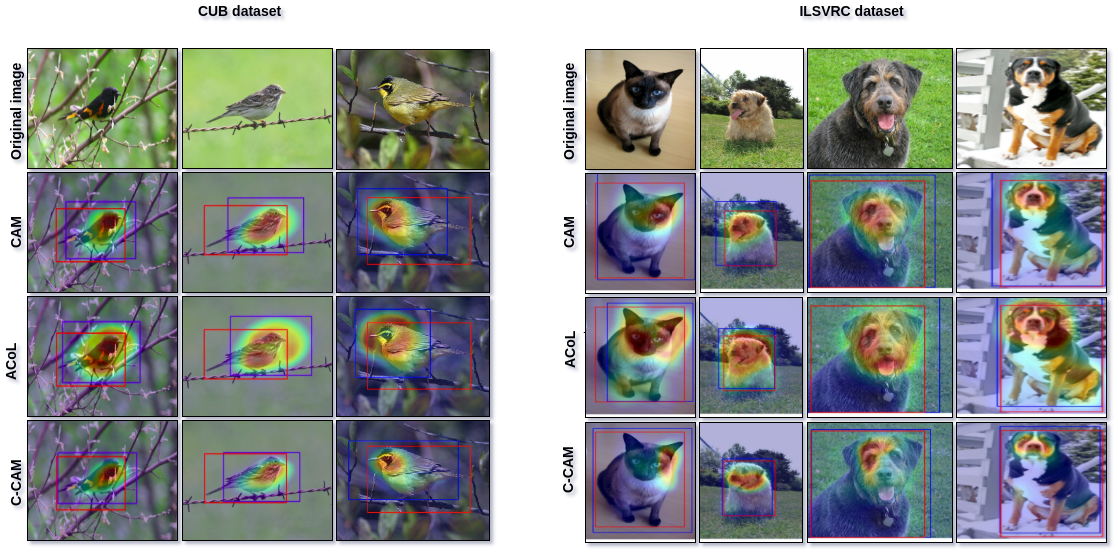}}
\caption{Comparison of HCLNet with CAM and ACoL method. HCLNet better localize the object than CAM \citep{b2} and ACoL \citep{b1} approaches. (ground-truth bounding boxes are in red and the predicted bounding boxes are in blue)}
\label{fig:cams}
\end{figure*}

\subsubsection{CUB-200-2011 dataset}
\textbf{Classification}: Table \ref{tab:1} shows the results of CAM, ACoL, SPG, ADL and the proposed method using VGGnet as backbone. Table \ref{tab:1} reports the Top-1 and Top-5 error on the CUB-200-2011 validation set. The proposed method achieved 22.11\%/6.44\% top-1/top-5 cls-err. This is better than Learning Deep Features (CAM) approach by 1.29\%/1.06\%. HCLNet has acheived a new state of the art for classification. It also performs 6.59\% lower than the baseline Adversarial complementary learning method (ACoL), and 12.62\% lower than Attention-based Dropout Layer (ADL) approach. The method has achieved these results because C-CAM does not erase any part from the CAM like previous methods.

\textbf{Localization}: On the CUB-200-2011 test set, HCLNet with addition strategy reports 48.95\%/41.06\% that is 6.9\%/6.78\% lower than Learning Deep Features (CAM) approach. It reports a margin of 5.13\%/2.43\% compared to  the baseline Adversarial complementary learning method (ACoL), and 2.12\%/1.09\% than Self-produced Guidance approach (SPG). ADL shows better results for localization, i.e a gap of 1.31\%. However, our method surpasses ADL in classification accuracy. 

The results of HCLNet with l1-norm strategy are 49.07\%/41.15\%. Compared to the other approaches, HCLNet is 5.9\%/6.68\% lower than Learning Deep Features (CAM) approach. It reports 5.01\%/2.33\% lower than Adversarial complementary learning method (ACoL), and 2\%/0.99\% lower than Self-produced Guidance approach(SPG). With this strategy, ADL still the best method for localization. 

Fig. \ref{fig:cams} shows some bounding boxes generated by HCLNet and compares them to CAM and ACoL approaches. HCLNet gives better localization and generates larger bounding boxes.

\textbf{CorLoc metric}: CorLoc can reflect the real localization performance as it does not take into account the classification accuracy. As shown in Tab. \ref{tab:corloc}, using VGG architecture as a backbone, HCLNet outperforms the baseline ACoL with a large margin of 6.06\% when both addition and $l_{1}$norm strategies are used. HCLNet outperforms the other state-of-the-art methods except for ADL where there is a difference of 12.72\%. ADL without considering the classification accuracy of the model localize better the objects. ADL uses a module where the most discriminative part is erased and an attention mechanism is used to reward this discriminative part. Changing the architecture of the network by adding an attention mechanism helpes the network to better localize the object, and hence gives better localization performance. This gives an intuition about the erasing method that detariorate the classification accuracy. And the attention mechanism that better localize the object.    

\subsubsection{ILSVRC dataset}
Due to the lack of training details i.e batch-size and number of epochs for each method are not given by the corresponding articles, we retrain \cite{b2,b1} with batch-size=32 for 10 epochs for fair comparison.

\textbf{Classification}: Tab. \ref{tab:2} shows the results of CAM and ACoL compared to HCLNet method with a VGGnet architecture as backbone on ILSVRC dataset. The proposed method achieved 21.51\% top-1 cls-err. It is 2.26\% lower top-1 cls-err than Learning Deep Features (CAM) approach and 4.0\% lower top-1 cls-err than the baseline Adversarial complementary learning method (ACoL). This further proves that C-CAM improved classification accuracy without using any extra parameters.

\textbf{Localization}: HCLNet with addition strategy and $l_{1}$norm strategy outperforms both CAM and ACoL methods with 1.58\% and 2.48\% respectively. 

\begin{table}[t]
\centering
 \begin{tabular}{|c|c|} 
 \hline
 Method & CorLoc \\ 
 \hline
 VGGnet-CAM \cite{b2} & 44.0 \\ 
 VGGnet-ACoL \cite{b1} & 45.9 \\
 VGGnet-SPG \cite{b5} & 41.1 \\
 VGGnet-ADL \cite{b4} & \textbf{25.22} \\
 HCLNet (addition) & 37.94 \\ 
 HCLNet ($l_{1}$-norm) & 37.94 \\
 \hline
 \end{tabular}
 \caption{CorLoc metric on the CUB-200-2011 test set}
 \label{tab:corloc}
\end{table}
\begin{table}[t]
\centering
 \begin{tabular}{|c|c|c|c|c|c|} 
 \hline
 \multicolumn{2}{|c|}{Method} &  \multicolumn{2}{c|}{cls-err} &  \multicolumn{2}{c|}{loc-err} \\ 
 \hline\hline
 CAM type & Fusion strategy & top1 & top5 & top1 & top5 \\ 
 \hline
 Threshold & Max  & 28.1 & - & 54.08 & 43.49 \\
 C-CAM & Max & 21.51 & 6.73 & 51.35 & 44.39 \\
 Threshold & addition & 28.1 & - & 51.54 & 43.03 \\
 Threshold & $l_{1}norm$ & 28.1 & - & 51.55 & 43.03 \\
 C-CAM & addition & \textbf{22.11} & \textbf{6.44} & 49.57 & 41.23 \\ 
 C-CAM & $l_{1}$-norm & \textbf{22.11} & \textbf{6.44} & \textbf{49.50} & \textbf{41.20} \\
 \hline
 \end{tabular}
 \caption{The effect of C-CAM and fusion strategy on the CUB-200-2011 test set.}
 \label{tab:ablation}
\end{table}
\subsection{Ablation study}
In this section, we show the effect of using C-CAM instead of a threshold. We also demonstrate the benefit of using a different fusion strategy on CUB-200-2011 using VGGnet. As shown in Tab. \ref{tab:ablation}, using only C-CAM with the max fusion strategy \cite{b1}, the cls-err is reduced by 6.59\%. The most discriminative part that helps the CNN to best classify the object in the image is not removed. This explain a higher classifcation accuracy. For loc-err using only C-CAM,  the error is decreased by 2.73\%/0.9\% top-1/top-5. A thresholded CAM + the addition fusion strategy improved the localization accracy by 2.54\%/0.46\% and the thresholded CAM + $l_{1}norm$ improved it by 2.53\%/0.46\%. As the fusion strategy is involved only in the localization process, there is no improvement in the classification accuracy. Addition strategy and $l_{1}norm$ strategy take into account all pixels of CAMs generated from the two branches. This is not the case of ACoL where they take only the max value of one of the two maps.

\section{Conclusion}
\label{sec:conclusion}
In this paper, we have presented a simple method to improve the classification and the localization in weakly supervised manner. We proposed to use a complementary map to detect the other features of the object. HCLNet compared to previous methods does not remove some parts of the image, and  then does not need any extra hyper-parameter. Furthermore, it improved both classification and localization of the baseline with less parameters. Two fusion strategies; the addition strategy and the $l_{1}$-norm strategy, have also been introduced. These two strategies fuse different class activation maps to generate the final object location. The experiments have shown that our method surpass previous state of the arts  approaches. This proposed method may find potential applications in  some area such as medical image processing, where an expert is needed to annotate images. 


\bibliography{hclnet}

\end{document}